\documentclass[runningheads]{llncs}
\usepackage{amsmath}
\usepackage{amssymb}
\usepackage{graphicx}
\usepackage{hyperref}
\usepackage{xcolor}

\usepackage{siunitx}
\usepackage{multirow}
\begin{document}
\title{Solving the Traveling Salesperson Problem with Precedence Constraints by Deep Reinforcement Learning}
\titlerunning{Solving TSPPC by DRL}
%
\author{Christian Löwens \and Inaam Ashraf \and Alexander Gembus \and Genesis Cuizon \and Jonas Falkner \and Lars Schmidt-Thieme}
\authorrunning{Löwens et al.}
%
\institute{University of Hildesheim, Germany,
\url{https://uni-hildesheim.de} 
\email{\{loewensc,ashraf,gembus,cuizon\}@uni-hildesheim.de}
\email{\{falkner,schmidt-thieme\}@ismll.uni-hildesheim.de}}
\maketitle              
\begin{abstract}

This work presents solutions to the Traveling Salesperson Problem with precedence constraints (TSPPC) using Deep Reinforcement Learning (DRL) by adapting recent approaches that work well for regular TSPs. Common to these approaches is the use of graph models based on multi-head attention (MHA) layers. One idea for solving the pickup and delivery problem (PDP) is using heterogeneous attentions to embed the different possible roles each node can take. In this work, we generalize this concept of heterogeneous attentions to the TSPPC. Furthermore, we adapt recent ideas to sparsify attentions for better scalability. Overall, we contribute to the research community through the application and evaluation of recent DRL methods in solving the TSPPC. Our code is available at \url{https://github.com/christianll9/tsppc-drl}.

\end{abstract}
\section{Introduction}

\noindent The Traveling Salesperson Problem (TSP) is an NP-hard problem. Many practically relevant operations research problems are formulated as variants of the TSP. In this work, we focus on the TSP with precedence constraints (TSPPC), a variation of the TSP that enforces special ordering constraints, i.e., node $i$ has to be visited before node $j$. Similar to the regular TSP, the TSPPC can be applied to many practically relevant problems, such as scheduling, routing, process sequencing, etc.

There has been a great deal of attention to the TSP and also work on the TSPPC, discussed in more detail in section \ref{sec:relatedWork}. In this work, we solve the TSPPC with Deep Reinforcement Learning (DRL) methods. Despite a large number of contributions to the TSP and the rising number of work that applies DRL methods to combinatorial optimization problems, to the best of our knowledge, there do not exist any DRL approaches to the TSPCC yet. Due to the NP-hard nature of the problem, it is not feasible to obtain optimal solutions within a reasonable computational time for large-size problems. Therefore, applying DRL is desirable, as Machine Learning models offer a fast inference time in addition to the ability to generalize and being used on different problem settings.

Building on the approach of Li et al. \cite{Li2021heterogeneous} we adapt their model to cope with precedence constraints. For that, we modify their heterogeneous attention layers to deal with precedence constraints as well as chains of precedence constraints. This enables the model to learn the constraints intrinsically. Furthermore, we sparsify the attentions and show that this not only leads to a better run time performance but also increases the overall performance of the model (cf. section \ref{sec:methodology}).

The rest of the paper is structured as follows. Section \ref{sec:relatedWork} highlights some of the most important related work. In section \ref{sec:problemSetting}, the problem setting is defined, while section \ref{sec:methodology} describes our methodology. Our experimental setup and results are presented in sections \ref{sec:experiments} and \ref{sec:results}, respectively. In section \ref{sec:Conclusion}, we end with some concluding remarks and thoughts on future work.

\section{Related Work}
\label{sec:relatedWork}
The TSPPC has been dealt with in a more generalized form in the field of operations research. It is termed Sequential Ordering Problem (SOP), which is another name for the asymmetric TSPPC. The SOP was initially presented as an operations research problem by \cite{escudero1988inexact}, who proposed a heuristic method to solve it. The first work to introduce an algorithm to the SOP for exact solutions \cite{ascheuer2000branch} formulates it as a mixed-integer linear programming (MILP) problem and uses a branch-and-cut algorithm. Later several exact branch-and-bound algorithms were proposed \cite{karan2011branch,mojana2012branch,shobaki2015exact}, improving the results of \cite{ascheuer2000branch}. In \cite{jamal2017solving} the authors improve on their previous work \cite{shobaki2015exact} by enhancing the lower bound method for the branch-and-bound approach. One of the best heuristic solvers for the traveling salesman problem is the LKH-3 algorithm \cite{helsgaun2017extension}. It includes an extension for solving SOPs.

\cite{bello2017neural} produced one of the significant works in solving combinatorial optimization problems using reinforcement learning. They used the Pointer Network of \cite{vinyals2015ptrnet} that consists of LSTM-based encoder-decoder architecture and applied an actor-critic algorithm. They achieve better results in comparison to the supervised learning of \cite{vinyals2015ptrnet} and the heuristics library OR-Tools \cite{vrp2016google}. Kool et al. \cite{kool2019attention} adapted the Transformer Model of \cite{Vaswani2017attention} to solve routing problems. They trained the model using the REINFORCE algorithm with a greedy rollout baseline and outperformed several TSP and VRP models, including \cite{bello2017neural}. 

\cite{Bdeir2021} and \cite{Falkner2020LearningTS} adapt the model from \cite{kool2019attention} to improve the performance on the CVRP and the CVRP-TW respectively  by making the feature embeddings more informative. \cite{lu2019learning} went one step ahead and utilized a consecutive improvement approach using a model partially based on the Transformer Model of \cite{Vaswani2017attention}. They developed a DRL-based controller that iteratively refines a random initial solution with an improvement operator. Their model outperformed \cite{kool2019attention} and other state-of-the-art models.

\cite{ma2019combinatorial} introduced Graph Pointer Networks (GPN) based on the Pointer Network of \cite{bello2017neural}. They combined it with a hierarchical policy gradient algorithm to achieve new state-of-the-art results. Although their model lags behind the attention model of \cite{kool2019attention} for small-scale TSPs, it outperforms every model as the scale increases. Moreover, they also conducted experiments for TSP with time window constraints showing new state-of-the-art results. 

\cite{xin2021multi} presented a Multi-Decoder Attention Model (MDAM) based on \cite{kool2019attention}. Instead of focusing on only one policy, their model learns multiple diverse policies and then utilizes a special beam search to pick the best of them. They also introduced an Embedding Glimpse layer to add more embedding information and thus improve each policy. \cite{Thyssens2022} solved the CVRP in a supervised fashion and outperforms \cite{kool2019attention} and \cite{xin2021multi} for fixed vehicle costs. 

Recently, \cite{Li2021heterogeneous} developed a DRL model using heterogeneous attentions for the pickup and delivery problem (PDP), which is a special case of the TSPPC \cite{dumitrescu2010traveling}. Here, every node is either a pickup or a delivery node, and every node is part of exactly one precedence constraint. Their model builds extensively on the attention model of \cite{kool2019attention} and also uses a greedy rollout baseline together with the REINFORCE algorithm. They conducted experiments using randomly generated data and achieved a smaller total tour length than \cite{kool2019attention} and OR-Tools.

\section{Problem Setting} \label{sec:problemSetting}
The TSP, at its core, is concerned with finding the shortest route between a given set of nodes $X$ while visiting each node only once and returning to the starting node. In addition, with precedence constraints, the starting point is prescribed. Furthermore, each route has to satisfy given precedence constraints. These constraints generally state something comparable to a visiting order. One node can be subject to many constraints. For example, $i$ has to be visited, before $j, k$ and $l$. The distance from node $i$ to $j$ is given by the distance matrix $D$. We want to find an optimal permutation $\sigma$ over the nodes, such that the total travel length $L$ is minimal. Adapted from \cite{ma2019combinatorial}, we can formulate a TSP with $n$ nodes:
\begin{equation}
\begin{gathered}
    \min L(\sigma,D)=\min\sum^n_{t=1}{D_{\sigma_t,\sigma_{t+1}}}\\
    \sigma_1 = \sigma_{n+1}\\
    \sigma_t \in \{1, ..., n\}\\
    \sigma_t \neq \sigma_{t'} \quad \forall\ t \neq t'
\end{gathered}
\end{equation}
$\tau(i)$ returns the ordering of node $i$ in the sequence according to the permutation $\sigma$, so that $i = \sigma_{\tau(i)}$. $P$ is the set of all precedence constraints. If $i$ has to precede $j$ this is represented by the pair $(i,j)$. Following this notation we can model the precedence constraints like:
\begin{equation}
    \tau(i) < \tau(j) \quad \forall\ (i,j) \in P
\end{equation}

\section{Methodology} \label{sec:methodology}
The proposed model builds on the work of \cite{kool2019attention} and \cite{Li2021heterogeneous}. We modify their Transformer model by restricting specific attentions. Similar to \cite{kool2019attention}, each problem instance $s$ can be considered as a graph with $n$ nodes (see section \ref{sec:problemSetting}) having features $x_i$, where $x_i$ are the 2D coordinates. Our graph would be fully connected in case of the unconstrained TSP. However, for the TSPPC, we use restricted attentions to let the model learn precedence constraints intrinsically \cite{Li2021heterogeneous}. Given a problem instance $s$, we sample the solution $\sigma$  from a stochastic policy $p(\sigma|s)$ determined by our Transformer model \cite{kool2019attention}:
\begin{equation}
    p_\theta(\sigma|s) = \prod_{t=1}^n p_\theta(\sigma_t|s, \sigma_{1:t-1})
\end{equation}
where the parameters $\theta$ describe the model. The encoder generates latent embeddings for all nodes using the coordinates as input features. These embeddings along with context are fed in to the decoder. The decoder works iteratively by decoding one state at a time to build the tour $\sigma$.

\subsection{Encoder}
The 2D input $x_i$ is embedded as $d_h$-dimensional vector $h_i^{(0)} = W^x x_i + b^x$, where $d_h=128$. Then, multiple stacked MHA layers aggregate the embeddings. The output of the final layer $h_i^{N}$ is used to compute a graph embedding $\bar{h}^{N} = \frac{1}{n} \sum_{i=1}^n h_i^{N}$, which can be interpreted as context.

\subsubsection{Heterogeneous attentions}
\label{sec:sub:heterogeneousAttentions}
In order to solve the PDP, the concept of heterogeneous attentions was introduced by \cite{Li2021heterogeneous}. In addition to the attentions from $n$ nodes to $n$ nodes, they introduce attentions (1) from every pickup/delivery node to its corresponding delivery/pickup node, (2) from every pickup/delivery node to all pickup nodes, and (3) from every pickup/delivery node to all delivery nodes. 

In the TSPPC, one node could have the role of a delivery of multiple pickups or a delivery and a pickup simultaneously. These are the cases that cannot occur in the PDP and therefore are not handled by \cite{Li2021heterogeneous}. To generalize their model to the TSPPC, we speak therefore of predecessors instead of pickups and successors instead of deliveries. Additionally, we restrict certain attentions from predecessors to successors ($ps$) and from successors to predecessors ($sp$).

The TSPPC can also include chains of precedence constraints, which is another aspect that makes the TSPPC different from the PDP. For example, we could model a problem as follows: Node C can only be visited after node B, which itself can only be visited after node A. All three nodes (A, B, and C) build a chain. So far, there are no specific attentions from C to A or vice versa. Thus, we add a third kind of heterogeneous attentions to our model, where we restrict attentions from and to all members ($mm$) of a chain of precedence constraints. We define $M$ as the set of all chains, where every node can be a member of a maximum of one chain. The function $chain(i)$ returns the chain of node $i$. Figure \ref{fig:het-attentions} illustrates the different heterogeneous attentions within the encoder.

\subsubsection{Sparse attentions}
\label{sec:sub:sparseAttentions}
\begin{figure}
    \centering
    \includegraphics[width=0.8\textwidth]{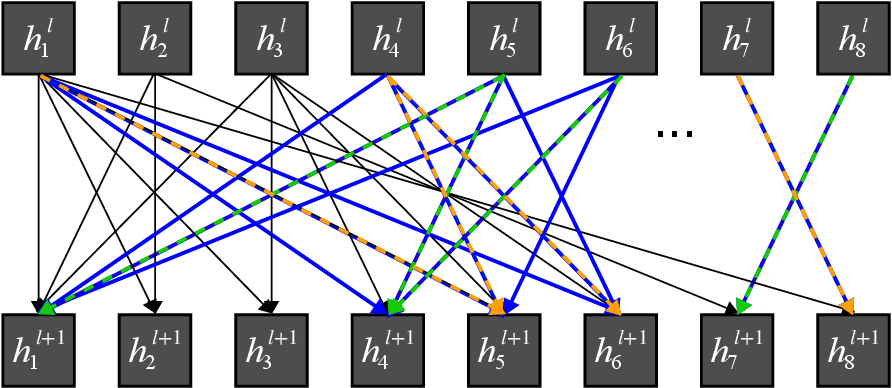}
    \caption{Heterogeneous attentions between two consecutive multi-head attention layers within the encoder. Black lines show attentions from neighbor-to-neighbor ($nn$) (for illustration reasons, only the first three nodes were used). Orange lines show attentions from predecessors to successors ($ps$) and green lines from successors to predecessors ($sp$). Blue lines show attentions to all other members of the same constraint group ($mm$).}
    \label{fig:het-attentions}
\end{figure}

Sparsifying a graph can improve the run time performance for large-scale problems. To achieve this, we restrict the attentions between all nodes so that node $i \in X$ can only reach its neighborhood $N_i$. This adds a forth kind of heterogeneous attentions to the model, namely attentions from neighbors to neighbors ($nn$). The black lines in figure \ref{fig:het-attentions} illustrate the attentions between neighbors for the first three nodes.

The neighborhood is calculated using two different approaches. First, we mask every attention between nodes, where the euclidean distance is larger than a fixed value $d_t$. The second approach uses the euclidean $k$-NN algorithm. Here, we mask all attentions to nodes that do not belong to the $k$ nearest neighbors. We discuss both approaches in section \ref{sec:experiments}.
\subsubsection{Formalization}
In total, we use four kinds of heterogeneous attentions $U=\{nn, ps, sp, mm\}$ within each layer $l\in \{1,...,N\}$ of our encoder. The kind of attention $u\in U$ is characterized by its own trainable weights for queries, keys and values ($W^{Q_{u}}, W^{K_{u}}, W^{V_{u}}$). Following loosely the notation of \cite{kool2019attention}, we can calculate the values for every node $i\in\{1,\dots,n\}$ and every layer $l$:
\begin{equation}
    q_i^{u} = W^{Q_{u}} h_i,\quad k_i^{u} = W^{K_{u}} h_{i},\quad v_i^{u} = W^{V_{u}} h_{i}
\end{equation}
When calculating the compatibilities $c_{ij}^{u}$, we mask some with $-\infty$ to restrict the corresponding attention:
\begin{align}
    c_{ij}^{nn}&= 
\begin{cases}
    \frac{{q_i^{nn}}^\intercal\ k_j^{nn}}{\sqrt{d_k}},& \text{if $j \in N_i$} \\
    -\infty,              & \text{otherwise}
\end{cases}\\
    c_{ij}^{ps}&= 
\begin{cases}
    \frac{{q_i^{ps}}^\intercal\ k_j^{ps}}{\sqrt{d_k}},& \text{if } (i,j) \in P\\
    -\infty,              & \text{otherwise}
\end{cases}\\
c_{ij}^{sp}&= 
\begin{cases}
\frac{{q_i^{sp}}^\intercal\ k_j^{sp}}{\sqrt{d_k}},& \text{if } (j,i) \in P\\
-\infty,              & \text{otherwise}
\end{cases}\\
c_{ij}^{mm}&= 
\begin{cases}
\frac{{q_i^{mm}}^\intercal\ k_j^{mm}}{\sqrt{d_k}},& \text{if}\ chain(i)=chain(j) \wedge\ chain(i)\in M\\
-\infty,              & \text{otherwise}
\end{cases}
\end{align}
To get the the attention weights $a_{ij}^{u}$ we apply the softmax function on the compatibilities.
\begin{equation}
    a_{ij}^{u} = \mathrm{softmax}(c_{i}^{u})_j
\end{equation}
The attention weights $a_{ij}^{u}$ are multiplied with values $v_j^{u}$ and then added to get the embeddings for each head as:

\begin{equation}
    h_i' = \sum_{u\in U}{\sum_{j\in n}{a_{ij}^{u} v_j^{u}}}
\end{equation}
We summarize the aforementioned calculations as applying the multi-head attention function MHA with $B=8$ heads on the embeddings $h$:
\begin{equation}
    \mathrm{MHA}(h)_i = \sum_{b=1}^B{W^O_b h_{ib}'}
\end{equation}
where $W^O$ is a trainable weight matrix.
Overall, each attention layer $l$ consists of a MHA function, a feedforward layer, skipped connections and a batch normalization (BN) function, which is applied twice to calculate the embedding $h_i^{l}$ of node $i$:
\begin{align}
    \hat{h}_i^l &= \mathrm{BN}^l(h_i^{(l-1)} + \mathrm{MHA}^l(h^{(l-1)})_i)\\
    h_i^{l} &= \mathrm{BN}^{l} (\hat{h}_i^l + \mathrm{FF}^l(\hat{h}_i^l))
\end{align}
We follow \cite{kool2019attention} and use the same graph embedding as the mean of all final node embeddings, which is used by the decoder for context.
\begin{equation}
    \bar{h}^N = \frac{1}{n}\sum_{i=1}^n{h_i^N}
\end{equation}

\subsection{Decoder}

Our decoder is similar to \cite{Li2021heterogeneous}, where it generates a probability vector based on the graph and node embeddings from the encoder. At the beginning, context $h^c$ consisting of graph embedding and last node embedding is required:
\begin{equation}
    h^c = Concat(\bar{h}^N, h_{\sigma_{t-1}}^N)
\end{equation}
Similarly, the glimpse $h^g = \mathrm{MHA}(W_g^Q h^c, W_g^K h^N, W_g^V h^N)$ is used for information aggregation. With the values $q_{(c)} = W^Q h^g$ and $k_i = W^K h_i^N$, we can compute the compatibility as:
\begin{equation}
    \hat{h}^t = C\cdot\tanh(h^t),
\end{equation} 
where
\begin{equation}
    h_i^t = 
    \begin{cases}
        \frac{q_{(c)}^T k_i}{\sqrt{d_k}},& \text{if } i \notin \sigma_{1:t-1} \wedge pred(i)\subseteq \sigma_{1:t-1} \\
        -\infty,              & \text{otherwise}
        \end{cases}
\end{equation}
Here, $pred(i)$ is a function returning the set of predecessors of node $i$. All visited nodes and all successors are masked until their corresponding predecessors are visited. $C$ is a hyperparameter used for clipping and is set to 10. Finally, we chose the next node to be visited based on the probability vector calculated as:
\begin{equation}
    p(\sigma_t|X, L_{t-1}) = \text{softmax}(\hat{h}^t)
\end{equation}
Similar to \cite{Li2021heterogeneous}, we optimize the loss $\mathcal{L}$ using gradient descent. We use the REINFORCE \cite{williams1992reinforce} algorithm with the greedy rollout \cite{kool2019attention} baseline $b(s)$:

\begin{equation}
    \nabla \mathcal{L} (\theta | s) = \mathbb{E}_{p_{\theta} (\sigma | s)} [(L(\sigma) - b(s)) \nabla \text{ log } p_{\theta} (\sigma | s)]
\end{equation}

\section{Experiments} \label{sec:experiments}

In order to fairly compare whether the models could generalize well across smaller and larger instances than those seen in training, we set up controlled experiments with fixed configurations. 

\subsection{Training and Datasets}
We train on fixed graph sizes TSPPC20 ($n_{train}=20$), TSPPC50  ($n_{train}=50$), TSPPC100  ($n_{train}=100$), with $|P|=0.33n$ precedence constraints. Each TSPPC instance consists of $n$ nodes sampled uniformly in the unit square $ S = {[x_i]}^n_{i=1} $ and $ x_{i}$ $\epsilon$  $[0, 1]^2 $. Each model is trained for 150 epochs on approximately 100,000 TSPPC samples, which are randomly generated for each epoch with the batch sizes of 512 (TSPPC20, TSPPC50) and 256 (TSPPC100) and a validation size of approximately 10,000.

\subsection{Model adaptations and sparsification}
To evaluate the usefulness of our adaptations, we train different versions of our model. First, we use the original model by Kool et al. \cite{kool2019attention} and adapt only the masking within the decoder so that all tours are feasible solutions for the TSPPC. Additionally, we train a model with and without attentions to the precedence chain members ($mm$). All these three models are trained on a dense graph without sparsification (i.e., $d_t=\infty \wedge k=\infty$).

Moreover, we try two approaches of sparse attentions. First, by restricting every attention between nodes, where the euclidean distance is larger than a particular threshold value $d_t$. Threshold values are chosen to be $d_t\in\{0.3, 0.5, 0.7, 0.9\}$. The second approach uses the euclidean $k$-NN algorithm, whereby all attentions to nodes, which do not belong to the $k\in\{5,10,20\}$ nearest neighbors, are restricted.

\subsection{Evaluation and Baselines} 
All models are evaluated for node sizes $n\in\{20, 50, 100, 150, 200\}$. We compare them against the simple Nearest Neighbor heuristic and the LKH-3 algorithm.

LKH-3 is a powerful, near-optimal solver for the TSPPC. Other work in the field of Reinforcement Learning often uses this solver and reports its results as the best-found solution, but consistently presents slow inference times \cite{bello2017neural,kool2019attention,ma2019combinatorial}. Although not reported specifically, this suggests the use of LKH-3's standard parameter setting without variation. We analyzed the influence of LKH-3's \emph{maxtrials} parameter on its cost-effectiveness. It can be observed that the standard-setting of maxtrials = 10,000 is not the most efficient choice. Based on this finding, we report differently parametrized versions of LKH-3 in section \ref{sec:results}.

\section{Results} \label{sec:results}
\begin{table}[h]
    \centering
    \caption{Average tour length and run time in seconds evaluation of the best performing models vs. baselines of 1,000 TSPPC samples for the node sizes $n=20$ and $n=50$. Note that bold figures in each node size represent the lowest tour length among baselines and models.}
    \label{tab:summary}
    \begin{tabular}{|l|rrr|rrr|}
    \hline
    \multicolumn{1}{|c|}{\multirow{2}{*}{Method}} &
    \multicolumn{3}{c|}{$n=20$} & \multicolumn{3}{c|}{$n=50$} \\
      &
      \multicolumn{1}{c}{Obj.} & \multicolumn{1}{c}{~~~~Gap} & \multicolumn{1}{c|}{~~~~Time} &
      \multicolumn{1}{c}{Obj.} & \multicolumn{1}{c}{~~~~Gap} & \multicolumn{1}{c|}{~~~~Time}  \\ \hline

     {LKH-3 (maxtrials = 1)}
     & 4.48 & 3.23\% & 0.041 & 7.13 & 9.86\% & 0.078  \\
     {LKH-3 (maxtrials = 10)}
     & 4.36 & 0.46\% & 0.047 & 6.71 & 3.39\% & \multicolumn{1}{r|}{0.090}  \\
     {LKH-3 (maxtrials = $\SI{10}{k}$)}
     & {\bfseries 4.34} & {\bfseries 0.00\%} & 1.160 & {\bfseries  6.49} & {\bfseries 0.00\%} & \multicolumn{1}{r|}{4.007} \\
     {Nearest Neighbor}
     & 5.33 & 22.81\% & \multicolumn{1}{r|}{\bfseries0.001} & 8.61 & 32.67\% & \multicolumn{1}{r|}{\bfseries0.005} \\ \hline

     {Kool et al. \cite{kool2019attention} (greedy, masked)} &
     5.12 &~~~~17.97\%& \multicolumn{1}{r|}{~~~~0.007} & 8.41 &~~~~29.58\% & ~~~~0.021 \\

     {Ours (greedy, dense)} &
     4.65 & 7.14\% & \multicolumn{1}{r|}{0.008} & 7.76 & 19.57\% & 0.028 \\
     
     {Ours (greedy, $k=5$)} &
     4.61 & 6.22\% & \multicolumn{1}{r|}{\bfseries0.005} & 7.66  & 18.03\% & {\bfseries 0.012}\\

     {Ours (sampling = $\SI{1}{k}$, $k=5$)} &
     4.40 & 1.38\% & \multicolumn{1}{r|}{0.034} & 7.15 & 10.17\% &  0.085 \\

     {Ours (sampling = $\SI{10}{k}$, $k=5$)} &
     {\bfseries 4.38} & {\bfseries 0.92\%} & \multicolumn{1}{r|}{0.123} & {\bfseries 7.07} & {\bfseries 8.94\%} & 0.483 \\
     \hline
     
    \end{tabular}
\end{table}

\begin{table} 
    \centering
    \caption{  Average tour length (and run time in seconds) comparison of TSPPC20, TSPPC50, and TSPPC100 models vs. baselines on the evaluation of 1,000 TSPPC samples at each varying node sizes. The number of precedence constraints is fixed at $|P|=0.33n$. Note that bold figures in each node size represent the lowest tour length among baselines and models.}
    \label{tab:tourlen}
    \resizebox{\textwidth}{!}{%
    \begin{tabular}{|l|r@{\hskip 0.1in}r|r@{\hskip 0.1in}r|r@{\hskip 0.1in}r|r@{\hskip 0.1in}r|r@{\hskip 0.1in}r|}
    \hline
    \multicolumn{1}{|c|}{\multirow{2}{*}{Method}} &
    \multicolumn{2}{c|}{$n=20$} & \multicolumn{2}{c|}{$n=50$}
    & \multicolumn{2}{c|}{$n=100$} & \multicolumn{2}{c|}{$n=150$} & \multicolumn{2}{c|}{$n=200$} \\
      &
      \multicolumn{1}{c}{Obj.} & \multicolumn{1}{c|}{Time} &
      \multicolumn{1}{c}{Obj.} & \multicolumn{1}{c|}{Time} &
      \multicolumn{1}{c}{Obj.} & \multicolumn{1}{c|}{Time} &
      \multicolumn{1}{c}{Obj.} & \multicolumn{1}{c|}{Time} &
      \multicolumn{1}{c}{Obj.} & \multicolumn{1}{c|}{Time}  \\ \hline
    {\itshape Baselines} & & & & & & & & & & \\
   LKH-3 (maxtrials = 1)      & 4.48 & 0.041 & 7.13 & 0.078 & 10.49 & 0.096 & 13.27 & 0.175 & 15.75  & 0.274\\
   LKH-3 (maxtrials = 10)     & 4.36 & 0.047 & 6.71 & 0.090 & 9.57 & 0.126  & 12.03 & 0.204 & 14.15 & 0.321  \\
   LKH-3 (maxtrials = $\SI{10}{k}$) & {\bfseries 4.34} & 1.160 & {\bfseries  6.49} & 4.007 & {\bfseries 8.93}  & 12.160 & {\bfseries 11.99} & 22.928 & {\bfseries 12.42} & 36.086 \\
    Nearest Neighbor          & 5.33 & 0.001 & 8.61 & 0.005 & 12.31 & 0.022 & 15.16 & 0.065 & 17.54 & 0.124 \\
    & & & & & & & & & &  \\
           
    {\bfseries TSPPC20}  & & & & & & & & & &  \\ 
    {\itshape Greedy Evaluation} & & & & & & & & & & \\ 
    Kool et al. \cite{kool2019attention} (masked)   & 5.12 & 0.007 & 8.51 & 0.021 & 12.93 & 0.077 & 16.66 & 0.257 & 20.05 & 0.345 \\
    Ours (dense, without $mm$)                     &  4.63 & 0.008 & 7.99 & 0.026 & 13.50 & 0.093 & 19.72 & 0.267 & 26.38 & 0.444  \\
   Ours (dense)     & 4.65 & 0.008 & 7.92 & 0.028 & 13.43 & 0.101 & 18.85 & 0.285 & 24.18 & 0.505 \\
   Ours ($d_t=0.5$) & 4.63 & 0.008 & 7.80 & 0.023 & 12.46 & 0.079 & 16.91 & 0.240 & 20.99 & 0.364 \\
   Ours ($k=5$)     & 4.61 & 0.005 & 7.86 & 0.015 & 12.75 & 0.049 & 17.09 & 0.103 & 21.09 & 0.176 \\
   Ours ($k=20$)    & 4.61 & 0.006 & 10.35 & 0.020 & 26.37 & 0.057 & 40.00 & 0.120 & 53.29 & 0.193 \\
   {\itshape Sampling (1000)} & & & & & & & & & & \\ 
   Ours ($d_t=0.5$) & 4.41 & 0.041 & 7.32 & 0.087 & 13.18 & 0.246 & 20.07 & 0.507 & 27.32 & 0.858 \\
   Ours ($k=5$)    &  4.40 & 0.034 & 7.26 & 0.085 & 12.41 & 0.247 & 17.67 & 0.519 & 22.90 & 0.825 \\
   Ours ($k=20$)   & 4.41 & 0.027 & 9.56 & 0.086  & 29.33 & 0.246 & 50.82 & 0.482 & 72.01 & 0.781 \\
   {\itshape Sampling (5000)} & & & & & & & & & & \\ 
   Ours ($d_t=0.5$) & 4.39  & 0.066  & 7.18 & 0.232 & 12.83  & 0.894 & 19.07  & 1.741 & 25.23 & 3.119 \\ 
   Ours ($k=5$)  & {\bfseries 4.38} & 0.078  &  7.16 & 0.272 & 12.20 & 0.803 & 17.38 & 1.737 & 22.54 & 3.015 \\ 
   Ours ($k=20$)  & 4.39 & 0.063 & 9.27 & 0.254 & 28.53 & 0.852 & 49.76 & 1.784 &  70.68 & 3.127 \\
   & & & & & & & & & &  \\
   {\bfseries TSPPC50} & & & & & & & & & & \\
    {\itshape Greedy Evaluation} & & & & & & & & & & \\
    Kool et al. \cite{kool2019attention} (masked) & 5.27 & 0.007 & 8.41 & 0.021 & 12.45 & 0.077 & 15.70 & 0.257 & 18.44 & 0.345 \\
    Ours (dense, without $mm$) & 5.09 & 0.008 & 7.76 & 0.026 & 11.52 & 0.093 & 14.94 & 0.267 & 18.19 & 0.444 \\
    Ours (dense) & 5.00 & 0.008 & 7.76 & 0.028 & 11.50 & 0.101 & 14.85 & 0.285 & 17.99 & 0.505 \\
    Ours ($d_t=0.5$) & 5.00 & 0.008 & 7.70 & 0.023 & 11.46 & 0.079 & 14.85 & 0.240 & 18.10 & 0.364 \\
    Ours ($k=5$)  & 4.95 & 0.004 &  7.66 & 0.012 &  11.30 & 0.050 &  14.54 & 0.108 & 17.47  & 0.175\\
    Ours ($k=20$) & 5.02 & 0.005 &  7.76 & 0.019 &  11.71 & 0.062 &  15.65 & 0.122 & 19.48 & 0.202 \\     
    {\itshape Sampling (1000)}  & & & & & & & & & &  \\ 
    Ours ($d_t=0.5$) & 4.67 & 0.030 & 7.16 & 0.080  & 10.83 & 0.238 & 14.61 & 0.553 & 18.51 & 0.863 \\ 
    Ours ($k=5$)     & 4.64 & 0.027 & 7.15 & 0.085  & 10.83 & 0.234 & 14.60 & 0.531 & 18.41 & 0.845 \\ 
    Ours ($k=20$)    & 4.70 & 0.031 & 7.23 & 0.086  & 11.31 & 0.276 & 16.18 & 0.494 & 21.49 & 0.806 \\   
    {\itshape Sampling (5000)}  & & & & & & & & & &  \\ 
    Ours ($d_t=0.5$) & 4.62 &  0.059 &  7.14 &  0.268 & 10.79 & 0.819 &  14.58 & 1.720 &  18.53 & 3.001\\ 
    Ours ($k=5$)  & 4.61  & 0.073 &  {\bfseries 7.09} & 0.250 & 10.68 & 0.827 & 14.38 & 1.749  & 18.16 & 3.037 \\ 
    Ours ($k=20$)  & 4.66 & 0.069 & 7.17 & 0.246 & 11.14 & 0.804 & 15.92 & 1.694 &  21.16 & 3.256 \\

    & & & & & & & & & & \\
    {\bfseries TSPPC100} & & & & & & & & & & \\
    {\itshape Greedy Evaluation} & & & & & & & & & & \\
   Ours ($d_t=0.5$) & 5.18 & 0.004 &  7.87 & 0.020  & 11.32 & 0.072   &  14.22 & 0.159 &   16.91 & 0.299 \\
   Ours ($k=5$) & 5.28 &  0.004 & 8.01 & 0.015 & 11.41 & 0.047 & 14.45 & 0.106 & 17.26 & 0.175 \\
   Ours ($k=20$) & 5.35 &  0.006 & 7.94 & 0.020 &  11.30 & 0.058 & 14.27 & 0.123 &  17.02 & 0.194\\

   {\itshape Sampling (1000)} & & & & & & & & & & \\
   Ours ($d_t=0.5$) & 4.71 & 0.027 & 7.29 & 0.078 &  10.63 & 0.232 &  13.84 & 0.524 & 17.05 & 0.827 \\
    Ours ($k=5$) & 4.81 &  0.031 & 7.49 & 0.087 & 10.77 & 0.236 & 13.94 & 0.507& 17.13 & 0.815 \\
    Ours ($k=20$)  & 4.81 & 0.031 & 7.35 & 0.077 & 10.68 & 0.237 & 13.95 & 0.482 & 17.33 & 0.840 \\

    {\itshape Sampling (5000)} & & & & & & & & & & \\
    Ours ($d_t=0.5$) & 4.66 & 0.056 & 7.21 &  0.253 & {\bfseries 10.51} &  0.852 & {\bfseries 13.65} & 1.727 & {\bfseries 16.84} & 2.993 \\
    Ours ($k=5$)  & 4.76 & 0.057 & 7.41 & 0.264 & 10.67 & 0.792 & 13.79 & 1.724 & 16.94 & 2.950 \\
    Ours ($k=20$)  & 4.76 & 0.058 & 7.27 & 0.260 & 10.55 & 0.846 & 13.77 & 1.738 & 17.10  & 2.943 \\
    \hline
    \end{tabular}}
   \end{table}

   We compare our best performing models for the node sizes $n=20$ and $n=50$ with the aforementioned baselines in Table~\ref{tab:summary}. It is worth noting that the results of the sampling approaches come quite close to those of the LKH-3. The default configuration of LKH-3 (with maxtrials = 10,000) has undoubtedly the best performance in finding the shortest tour but also has the longest run time across all node sizes. Nevertheless, by using a different configuration, LKH-3 is still able to compete against all sampling approaches in tour length and run time. On the other hand, the Nearest Neighbor baseline gives the worst tour length performance but the fastest run times. Furthermore, we can see that using heterogeneous attentions can achieve better results compared to a masking approach based on the model by \cite{kool2019attention}.

   The sparse model (where $k=5$) outperforms its dense counterpart not only in terms of run time but also achieves a shorter average tour length. The reason for this might be that sparsification forces the model to search for optimal follow-up nodes in the proximity of the current node. The hypothesis that the optimal next node lies close to the current node intuitively makes sense. By enforcing this assumption through sparsification, the model focuses on the most promising next candidate nodes, thus achieving better performance.
    
   To evaluate the ability to generalize, we show the results of our models for different node sizes in Table~\ref{tab:tourlen}. Comparing TSPPC20, TSPPC50, and TSPPC100 models, the common observation is that models trained at a particular node size perform better when evaluated on the same node size they are trained on. Furthermore, TSPPC50 models tend to scale better when evaluated on larger node sizes (i.e., $n=100$, 150, and 200). Focusing on TSPPC50 models, the sparse $k$-NN $k=5$ model outperforms all other dense and sparse models in all node sizes they are evaluated on. Moreover, it is also the only model that beats the Nearest Neighbor baseline at $n=200$, indicating the scalability of this sparse model to larger node sizes. Note that the results for other model parameters can be found in the appendix table in section \ref{sec:Appendix}.

   During our experiments, we observed unusual behavior regarding the scalability of some of our models. For instance, TSPPC20 $k$-NN $k=20$ model in Table \ref{tab:tourlen} exhibits significantly worse results when scaled to higher node sizes. TSPPC20 with 20 nearest neighbors means that the model has access to all nodes of the graph during training, which essentially makes it a dense model. Thus, it should also scale in a similar way as the dense model. We hypothesize that the reason behind the poor scalability is the fact that the model is practically trained as a dense model and then evaluated with sparse attentions for larger node sizes. Hence, we conclude that a model needs to be trained on sparse attentions if it is to be evaluated on sparse attentions.
   
    While the $k$-NN $k=5$ model achieved the best performance for both, TSPPC20 and TSPPC50 models, we can see that the threshold-based sparse TSPPC100 model (where $d_t=0.5$) gives a slightly better tour length than the corresponding $k$-NN models. Therefore, we would say that no single sparse model works best for all node sizes. Hence, the sparsity level should be carefully selected since it is not a trivial task. Interestingly, we can see that the effect of sampling methods shrinks when increasing $n$ because the solution space grows exponentially. For $n=200$, the greedy approach already beats the sampling of 1000 tours.

      

   \section{Conclusion and Future Work}
   \label{sec:Conclusion}
   In this paper, we successfully deploy a DRL-based training method using a Transformer model to solve the TSPPC. Furthermore, we sparsify our attentions, achieving not only faster computation time but also a gain in performance. Our model achieves better results than the model from Kool et al. \cite{kool2019attention} with a simple masking adaptation for the TSPPC. However, the scalability of our model to very large sizes was lacking in our experiments, which can be an aspect to concentrate on in future work. 
   
   We analyze the LKH-3 heuristic algorithm and, against what most other literature does, also report results for non-standard LKH-3 settings. This shows that LKH-3, while still having outstanding performance, can be very fast in inference as well.
   
   It can be said that DRL methods applied to the TSPPC have shown promising results, notably when evaluated on the same problem size as seen in training. Also, it is scalable to different node sizes that are close to the original training node size.

    \subsubsection{Acknowledgment} This work was supported by the German Federal Ministry of Education and Research (BMBF) via the project "Learning to Optimize" (L2O) under the grant no. 01IS20013A.
   
 \clearpage

%
%
%
\bibliographystyle{splncs04}
%

\clearpage
   
\section{Appendix}
\label{sec:Appendix}

\begin{table}[h]
    \centering
    \caption{Average tour length (and run time in seconds) comparison of all TSPPC20 and TSPPC50 models vs. baselines evaluated on 1,000 TSPPC samples at each varying node sizes. The number of precedence constraints is fixed at $|P|=0.33n$. For a better comparison of the effects caused by using different model parameters, we show only the greedy evaluation.}
    \label{tab:allexp}
    \resizebox{\textwidth}{!}{%
        \begin{tabular}{|l|r@{\hskip 0.1in}r|r@{\hskip 0.1in}r|r@{\hskip 0.1in}r|r@{\hskip 0.1in}r|r@{\hskip 0.1in}r|}
            \hline
            \multicolumn{1}{|c|}{\multirow{2}{*}{Method}} &
            \multicolumn{2}{c|}{$n=20$} & \multicolumn{2}{c|}{$n=50$}
            & \multicolumn{2}{c|}{$n=100$} & \multicolumn{2}{c|}{$n=150$} & \multicolumn{2}{c|}{$n=200$} \\
              &
              \multicolumn{1}{c}{Obj.} & \multicolumn{1}{c|}{Time} &
              \multicolumn{1}{c}{Obj.} & \multicolumn{1}{c|}{Time} &
              \multicolumn{1}{c}{Obj.} & \multicolumn{1}{c|}{Time} &
              \multicolumn{1}{c}{Obj.} & \multicolumn{1}{c|}{Time} &
              \multicolumn{1}{c}{Obj.} & \multicolumn{1}{c|}{Time}  \\ \hline
    {\itshape Baselines}  & & & & & & & & & & \\
   LKH-3 (maxtrials = 1) & 4.48 & 0.041 & 7.13 & 0.078 & 10.49 & 0.096 & 13.27 & 0.175 & 15.75  & 0.274\\
   LKH-3 (maxtrials = 10) & 4.36 & 0.047 & 6.71 & 0.090 & 9.57 & 0.126  & 12.03 & 0.204 & 14.15 & 0.321  \\
   LKH-3 (maxtrials = $\SI{10}{k}$) & {\bfseries 4.34} & 1.160 & {\bfseries  6.49} & 4.007 & {\bfseries 8.93}  & 12.160 & {\bfseries 11.99} & 22.928 & {\bfseries 12.42} & 36.086 \\
    Nearest Neighbor & 5.33 & 0.001 & 8.61 & 0.005 & 12.31 & 0.022 & 15.16 & 0.065 & 17.54 & 0.124 \\
    & & & & & & & & & & \\
           
    {\bfseries TSPPC20}  & & & & & & & & & &\\ 
    {\itshape Greedy Evaluation} & & & & & & & & & & \\ 
    Kool et al. \cite{kool2019attention} (masked)   & 5.12 & 0.007 & 8.51 & 0.021 & 12.93 & 0.077 & 16.66 & 0.257 & 20.05 & 0.345 \\
    Ours (dense, without $mm$)                     &  4.63 & 0.008 & 7.99 & 0.026 & 13.50 & 0.093 & 19.72 & 0.267 & 26.38 & 0.444  \\
   Ours (dense)     & 4.65 & 0.008 & 7.92 & 0.028 & 13.43 & 0.101 & 18.85 & 0.285 & 24.18 & 0.505 \\
   Ours ($d_t=0.3$) & 4.63 & 0.006 & 7.88 & 0.189  & 13.20 & 0.062 & 19.05 & 0.178 & 24.75 & 0.303 \\
   Ours ($d_t=0.5$) & 4.63 & 0.008 & 7.80 & 0.023 & 12.46 & 0.079 & 16.91 & 0.240 & 20.99 & 0.364 \\
   Ours ($d_t=0.7$) & 4.64 & 0.008 & 7.96 & 0.026 & 13.19 & 0.089 & 18.28 & 0.312 & 23.20 & 0.445 \\
   Ours ($d_t=0.9$) & 4.63 & 0.008 & 7.92 & 0.028  & 13.08 & 0.095 & 17.98 & 0.277 & 22.50 & 0.500\\
   Ours ($k=5$)     & 4.61 & 0.005 & 7.86 & 0.015 & 12.75 & 0.049 & 17.09 & 0.103 & 21.09 & 0.176 \\
   Ours ($k=10$)   & 4.62 & 0.005 & 8.13 & 0.016  & 14.13 & 0.050 & 19.63 & 0.112 & 24.74 & 0.190 \\
   Ours ($k=20$)    & 4.61 & 0.006 & 10.35 & 0.020 & 26.37 & 0.057 & 40.00 & 0.120 & 53.29 & 0.193 \\
   & & & & & & & & & & \\
   {\bfseries TSPPC50} & & & & & & & & & &\\
   {\itshape Greedy Evaluation} & & & & & & & & & & \\
    Kool et al. \cite{kool2019attention} (masked)         & 5.27 & 0.007 & 8.41 & 0.021 & 12.45 & 0.077 & 15.70 & 0.257 & 18.44 & 0.345 \\
    Ours (dense, without $mm$)          & 5.09 & 0.008  & 7.76 & 0.026 & 11.52 & 0.093 & 14.94 & 0.267 & 18.19 & 0.444 \\
    Ours (dense)        & 5.00 & 0.008  &  7.76 & 0.028 & 11.50 & 0.101 & 14.85 & 0.285 & 17.99 & 0.505  \\
    Ours ($d_t=0.3$)& 4.97 & 0.006 & 7.66 & 0.188 & 11.37 & 0.062 & 14.76 & 0.178 & 17.96 & 0.302 \\
    Ours ($d_t=0.5$) & 5.00 & 0.008 & 7.70 & 0.023 & 11.46 & 0.079 & 14.85 & 0.240 & 18.10 & 0.364 \\
    Ours ($d_t=0.7$)& 5.02 & 0.008 & 7.72 & 0.026 & 11.47 & 0.089 & 14.97 & 0.312 & 18.28 & 0.445 \\
    Ours ($d_t=0.9$)& 5.01 & 0.008 & 7.69 & 0.028 & 11.39 & 0.095 & 14.66 & 0.277 & 17.68 & 0.500 \\
    Ours ($k=5$)  & 4.95 & 0.004 &  7.66 & 0.012 &  11.30 & 0.050 &  14.54 & 0.108 &  17.47  & 0.175\\
    Ours ($k=10$)  & 5.05 & 0.005 & 7.73 & 0.014 & 11.61 & 0.054 & 15.18 & 0.110 & 18.50 & 0.187\\
    Ours ($k=20$) & 5.02 & 0.005 &  7.76 & 0.019 &  11.71 & 0.062 &  15.65 & 0.122 & 19.48 & 0.202 \\
    \hline
    \end{tabular}}
   \end{table}

\end{document}